\documentclass[11pt]{article}

\usepackage{acl}

\usepackage{times}
\usepackage{latexsym}

\usepackage[T1]{fontenc}
 
\usepackage[utf8]{inputenc}
\usepackage{times}
\usepackage{enumitem}
\usepackage{latexsym}
\usepackage{microtype}
\usepackage{makecell}
\usepackage{hyperref}
\usepackage{graphicx}
\usepackage{multirow}
\usepackage{amsmath}
\usepackage{xcolor}
\usepackage{booktabs}
\usepackage{caption}
\usepackage{subcaption}
\usepackage{amsfonts}
\usepackage{tabularx}
\usepackage{rotating}
\usepackage[T1]{fontenc}
\usepackage[utf8]{inputenc}
\usepackage{microtype}
\usepackage{inconsolata}
\usepackage{graphicx}
\usepackage{amssymb}
\usepackage{algorithmicx}
\usepackage{algorithm}
\usepackage{algpseudocode}
\usepackage{amsmath}
\usepackage{dblfloatfix}  %
\usepackage{booktabs}
\usepackage{multirow}
\usepackage[table]{xcolor}
\usepackage{pifont}
\newcommand{\xmark}{\ding{55}}

\usepackage{subcaption}
\usepackage{subcaption} 
\algnewcommand\algorithmicendwhile{\textbf{end\ while}}
\algrenewtext{While}[1]{\algorithmicwhile\ #1\ \algorithmicdo}
\usepackage{caption}
\usepackage{xcolor}

\usepackage[table]{xcolor}

\definecolor{lightblue}{RGB}{203, 204, 255}
\definecolor{lightgray}{RGB}{229, 229, 229}

\title{Cross-lingual Representation Learning via Centroid Intervention Fusion}

\author{First Author \\
  Affiliation / Address line 1 \\
  Affiliation / Address line 2 \\
  Affiliation / Address line 3 \\
  \texttt{email@domain} \\\And
  Second Author \\
  Affiliation / Address line 1 \\
  Affiliation / Address line 2 \\
  Affiliation / Address line 3 \\
  \texttt{email@domain} \\}

\author{Wei Sun \and Marie-Francine Moens \\
        Department of Computer Science, KU Leuven \\
        Celestijnenlaan 200A 3001 Heverlee, Belgium \\
        \texttt{\{sun.wei, sien.moens\}@kuleuven.be}}

\begin{document}
\maketitle

\begin{abstract}
Large language models (LLMs) exhibit uneven multilingual performance, especially when dealing with low-resource languages.
Inference-time intervention offers a lightweight way to improve cross-lingual transfer by modifying the hidden states produced by the LLMs during the forward pass, without updating model parameters.
However, existing cross-lingual intervention methods typically learn separate projections from source to target languages, which limits scalability and prevents knowledge sharing across languages.
We propose \textbf{Centroid Intervention Fusion (CIF)}, a projection fusion framework that consolidates multiple multilingual intervention projections into a single language-shared operator.
Across multilingual commonsense reasoning, natural language inference, factual editing, and machine translation benchmarks, CIF outperforms the strongest prior pairwise intervention baseline by up to \textbf{+3.378 pp} on average across four model backbones, while supporting performance gains for low resource languages.
The code is available at \url{https://github.com/VRCMF/CIF.git}. 
\end{abstract}

\section{Introduction}

Large language models (LLMs) have shown strong multilingual abilities in many understanding and generation tasks~\citep{xue-etal-2021-mt5,lin-etal-2022-shot,muennighoff-etal-2023-crosslingual,ustun-etal-2024-aya,bai2023qwen,touvron2023llama2,jiang2023mistral,dubey2024llama3}.
However, their performance remains highly uneven across languages, with low-resource languages often lagging far behind high-resource ones~\citep{pires-etal-2019-multilingual,conneau2019cross,wu-dredze-2020-languages,petrov2023language}.
Existing remedies usually rely on continued pretraining or supervised fine-tuning with multilingual data~\citep{xue-etal-2021-mt5,muennighoff-etal-2023-crosslingual,ustun-etal-2024-aya}, which can be costly in both data and computation.

\begin{figure}[t]
    \centering
    \includegraphics[width=0.5\textwidth]{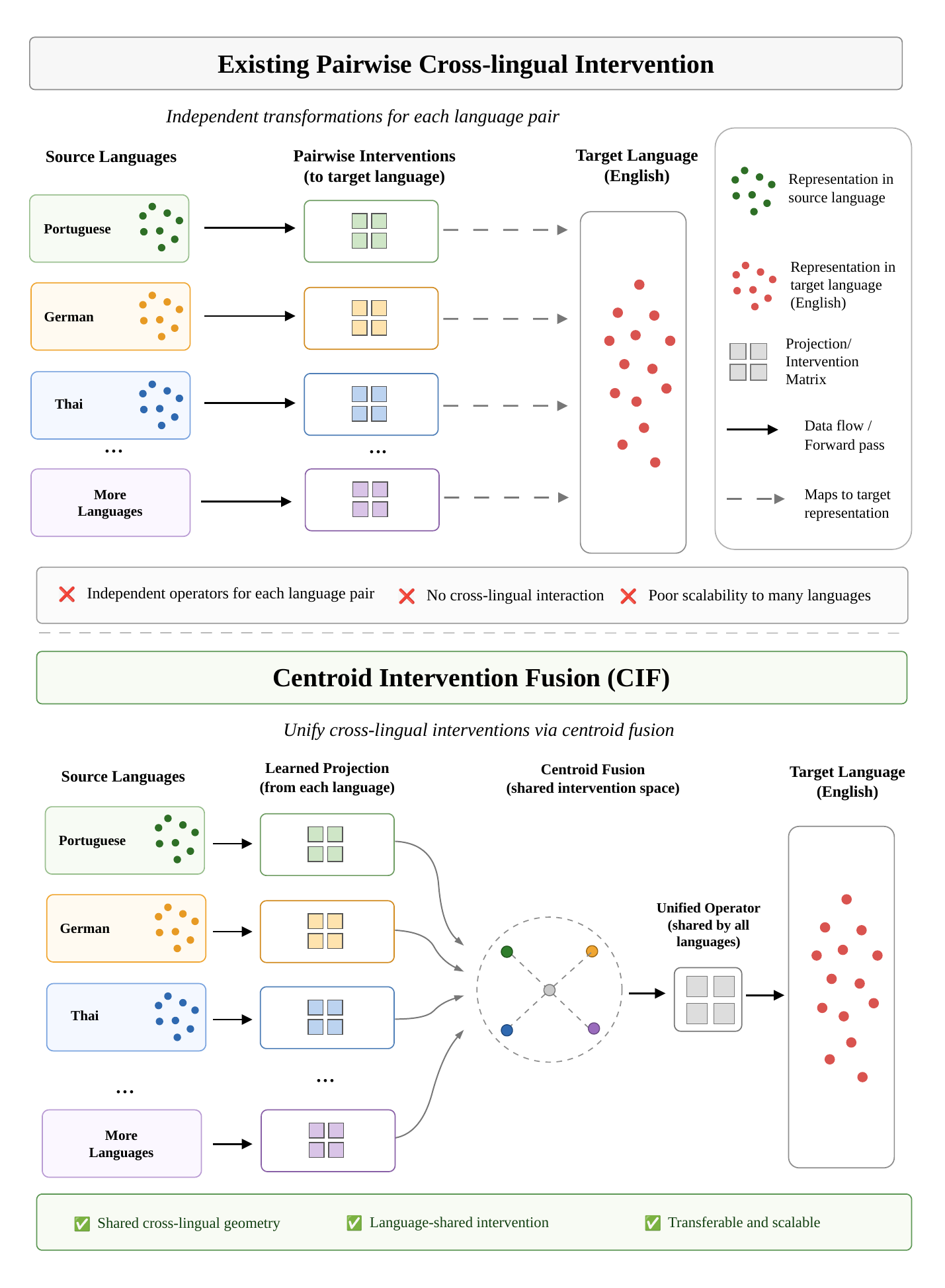}
    \caption{
    Comparison between existing pairwise cross-lingual intervention methods and CIF.
Existing methods use separate intervention operators for each language pair, while CIF unifies separate operators into a single language-shared operator via centroid fusion.
    }
    \label{fig:intro_comparison}
\end{figure}

Inference-time intervention offers a lightweight alternative by modifying the hidden states produced by the LLMs during the forward pass,
without updating model parameters~\citep{subramani-etal-2022-extracting,turner2023activation,zou2023representation,li2023inference,rimsky-etal-2024-steering,wang2025sadi}.
Recent multilingual intervention methods transform representations for different languages; for example, INCLINE learns a projection matrix for each source-target language pair~\citep{wang2025incline}.
Although effective, this pairwise design has two drawbacks: it requires a separate operator for every transfer representation, and it cannot explicitly share useful intervention knowledge across language pairs.
As shown in Figure~\ref{fig:intro_comparison}, our goal is to replace pairwise operators with one shared cross-lingual operator.

Our motivation is that cross-lingual models often organize different languages in partially shared semantic spaces~\citep{conneau2019cross,pires-etal-2019-multilingual,artetxe2019massively,feng2022language,zhao-etal-2021-inducing,tiyajamorn-etal-2021-language}.
This suggests that projection matrices learned from different language pairs may contain common cross-lingual representations, even though each language also has its own variation.
We hypothesize that these common representations can be estimated by fusing multiple pairwise projections into a single shared intervention operator.

We propose \textbf{Centroid Intervention Fusion (CIF)}, a simple projection-fusion method.
Given projection matrices from multiple language pairs, CIF first rewrites each matrix as a perturbation around the identity transformation.
It then estimates a robust centroid via centroid-guided trimming, where projections farthest from the centroid are discarded before averaging the remaining ones. Motivated by the observation that multilingual models may suffer from negative interference across languages~\cite{wang-etal-2020-negative}, this filtering step aims to prevent atypical language-pair projections from dominating the shared operator.

We evaluate CIF in downstream tasks that concern multilingual commonsense reasoning, natural language inference, factual editing, and machine translation, using the XCOPA, XStoryCloze, XWinograd, XCSR, XNLI, MzsRE, WMT23, and FLORES-101~\citep{ponti-etal-2020-xcopa,lin-etal-2022-shot,tikhonov-ryabinin-2021-heads,lin-etal-2021-common,conneau-etal-2018-xnli,wang-etal-2024-retrieval,kocmi-etal-2023-findings,goyal-etal-2022-flores} datasets.
Experiments with Qwen, BLOOMZ, LLaMA, and Mistral~\citep{bai2023qwen,muennighoff-etal-2023-crosslingual,touvron2023llama,touvron2023llama2,dubey2024llama3,jiang2023mistral} show that CIF consistently improves cross-lingual intervention performance.
Compared with the strongest pairwise baseline, CIF improves the average score across four model backbones by up to \textbf{3.378 pp}.

\noindent \textbf{Our contributions are:}
\begin{itemize}[leftmargin=*, noitemsep, topsep=0pt]
    \item We reveal the scalability and knowledge-sharing limitations of pairwise cross-lingual intervention.
    \item We propose \textbf{Centroid Intervention Fusion (CIF)}, which fuses multiple projection matrices into one robust language-shared intervention operator.
    \item We validate CIF across four LLM families, multiple model backbones, and diverse language families, showing consistent improvements over pairwise intervention.
\end{itemize}

\section{Related Work}

\paragraph{Inference-Time Intervention}

Inference-time intervention controls the behavior of large language models (LLMs) by modifying the hidden states produced by the LLMs during the forward pass
during inference, rather than updating model parameters~\citep{subramani-etal-2022-extracting,turner2023activation,zou2023representation}.
Such interventions have been used to steer model generations towards desired properties, including truthfulness, lower toxicity, and controllable attributes such as sentiment, topic, and factuality~\citep{li2023inference,turner2023activation,zou2023representation,rimsky-etal-2024-steering}.

Existing methods differ mainly in how the steering signal is constructed.
ITI~\citep{li2023inference} identifies attention heads associated with truthful behavior and edits their activations at inference time.
Activation Addition~\citep{turner2023activation} builds steering 
representations from contrastive prompts, while CAA~\citep{rimsky-etal-2024-steering} averages activation differences between positive and negative examples of a target behavior.
SADI~\citep{wang2025sadi} further makes the steering vector adaptive to input semantics.
These studies show that model behaviors can be controlled through representations in the activation space.
However, they do not explicitly study whether intervention representations learned in different languages share a transferable structure.

\begin{figure*}[t]
    \centering
    \includegraphics[width=\textwidth]{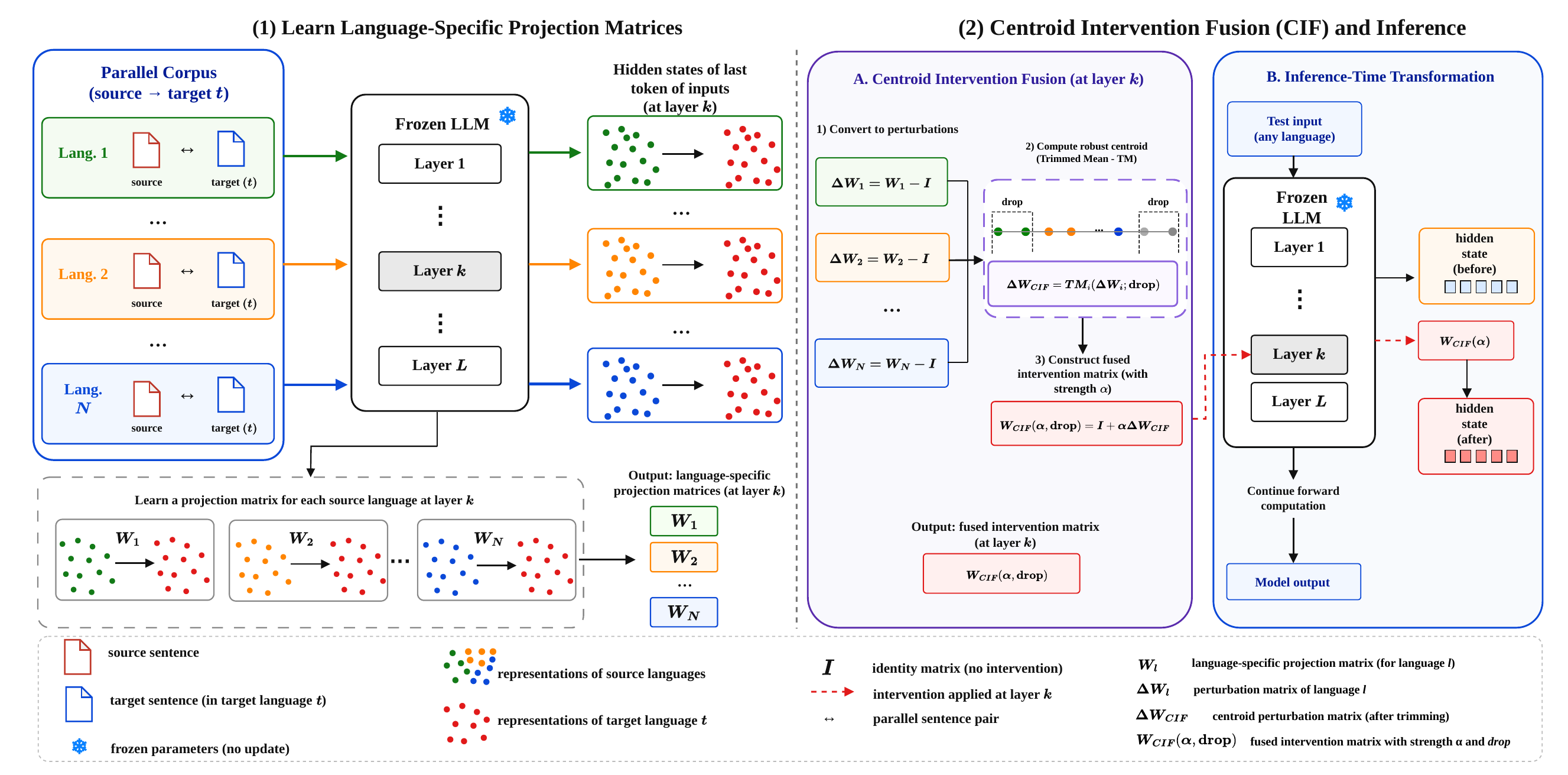}
    \caption{
    Overview of Centroid Intervention Fusion (CIF).
    CIF first learns language-specific projection matrices from multilingual parallel corpora and converts them into intervention perturbations around the identity transformation.
    It then computes a centroid perturbation in the projection weight space to construct a unified language-shared intervention matrix.
    During inference, the fused intervention matrix is applied to hidden representations of inputs to enable stable and transferable cross-lingual intervention across languages.
    }
    \label{fig:cif_framework}
\end{figure*}

\paragraph{Cross-Lingual Representation Alignment}

Cross-lingual representation learning studies how to represent inputs from different languages in a shared semantic space~\citep{conneau2019cross,artetxe2019massively,feng2022language}.
Multilingual pretrained models have been shown to support zero-shot transfer across languages, suggesting that they encode partially shared representations, although transfer quality can still vary across language pairs~\citep{pires-etal-2019-multilingual}.
Multilingual sentence encoders further demonstrate that shared representation spaces can benefit cross-lingual retrieval, semantic similarity, and cross-lingual transfer~\citep{artetxe2019massively,feng2022language,tiyajamorn-etal-2021-language}.

Earlier alignment methods map word embeddings from different languages into a common space with learned transformations~\citep{mikolov2013exploiting,artetxe2018robust}.
This idea has also been extended to contextual representations for cross-lingual transfer~\citep{schuster2019crosslingual}.
Recent work further seeks to reduce outlier signals and induce 
stronger language-invariant representations~\citep{zhao-etal-2021-inducing,tiyajamorn-etal-2021-language}.
Its findings suggest that multilingual models may contain shared geometric structure beyond language-specific representations.
Our work builds on this view but shifts the target from aligning representations to consolidating intervention operators learned from different languages.

\paragraph{Cross-lingual Intervention}

Recent work has begun to combine inference-time intervention with cross-lingual representation alignment.
INCLINE~\citep{wang2025incline} learns source-to-target alignment matrices from parallel sentences and applies them at inference time to transform representations from lower-performing languages towards higher-performing language spaces.
This shows that representation-level transformations can improve cross-lingual model behavior without changing model weights.
However, INCLINE mainly models cross-lingual intervention as independent pairwise transformations, which limits knowledge sharing across languages as the language set grows.
Our work instead studies whether such intervention operators can be consolidated across languages.

\section{Method}
\label{sec:method}

Figure~\ref{fig:cif_framework} illustrates \textbf{Centroid Intervention Fusion (CIF)}.
CIF converts a set of language-specific projection matrices into a single language-shared intervention matrix.
The motivation is that each source language may require a different correction toward an anchor-language representation space, while these corrections should still contain a shared cross-lingual representation.
CIF estimates this shared representation in three steps: first it learns one linear projection for each source language, then removes projection residuals that are far from the centroid, and finally applies the fused residual as an inference-time intervention.
The language model is kept frozen throughout, following prior work on activation-level steering \citep{subramani-etal-2022-extracting,turner2023activation}.
For clarity, we describe CIF at one layer $k$; the same procedure is applied independently to all layers.

\subsection{Learning Language-Specific Projection Matrices}
\label{sec:3.1}

Let $\mathcal{S}$ be the set of source languages and let $a$ be a shared anchor language, such as English.
For each source language $s \in \mathcal{S}$, we use a parallel corpus
\begin{equation}
\mathcal{D}_s
=
\{(x_i^s, x_i^a)\}_{i=1}^{N_s},
\end{equation}
where $x_i^s$ is a sentence in language $s$ and $x_i^a$ is its translation in the anchor language.
We feed both sentences into the frozen model and extract the last non-padding token representations at layer $k$:
\begin{equation}
\mathbf{h}_{i,k}^s \in \mathbb{R}^{d},
\qquad
\mathbf{h}_{i,k}^a \in \mathbb{R}^{d}.
\end{equation}

For each language $s$, we learn a linear projection matrix
$\mathbf{W}_{s,k} \in \mathbb{R}^{d \times d}$
that maps source-language hidden states to the anchor-language hidden-state space.
This follows the common view that cross-lingual representation spaces can be aligned with linear transformations \citep{mikolov2013exploiting,conneau2018word}.
The projection $\mathbf{W}_{s,k}^{*}$ is obtained by least squares:
\begin{equation}
\begin{aligned}
\arg\min_{\mathbf{W}_{s,k} \in \mathbb{R}^{d \times d}}
\frac{1}{N_s}
\sum_{i=1}^{N_s}
\Bigl\|
\mathbf{W}_{s,k}\mathbf{h}_{i,k}^{s}
-
\mathbf{h}_{i,k}^{a}
\Bigr\|_2^2 .
\end{aligned}
\end{equation}
This gives one language-specific projection matrix for each source language,
$\{\mathbf{W}_{s,k}^{*}\}_{s \in \mathcal{S}}$.

\subsection{Learning Language-Shared Projection via Centroid-Guided Intervention Fusion}

Directly averaging all projection matrices can be sensitive to languages whose learned projection contains a large language-specific correction.
CIF therefore fuses the residual intervention induced by each projection, rather than the full projection itself.
For each language, we rewrite the learned projection as a residual perturbation around the identity matrix:
\begin{equation}
\Delta \mathbf{W}_{s,k}
=
\mathbf{W}_{s,k}^{*}
-
\mathbf{I}_d .
\end{equation}
Here, $\mathbf{I}_d$ corresponds to no intervention, so
$\Delta \mathbf{W}_{s,k}$ captures only the change introduced by language $s$.

CIF then applies a centroid-based trimming step before averaging.
This follows the robust-estimation intuition behind trimmed aggregation, where outlying updates are removed before computing the mean \citep{huber2009robust,yin2018byzantine}.
We first compute the centroid of all residual perturbations:
\begin{equation}
\bar{\Delta \mathbf{W}}_{k}
=
\frac{1}{|\mathcal{S}|}
\sum_{s \in \mathcal{S}}
\Delta \mathbf{W}_{s,k}.
\end{equation}
Next, we measure how far each language-specific residual is from this centroid:
\begin{equation}
d_{s,k}
=
\left\|
\Delta \mathbf{W}_{s,k}
-
\bar{\Delta \mathbf{W}}_{k}
\right\|_F ,
\end{equation}
where $\|\cdot\|_F$ denotes the Frobenius norm, which measures the overall Euclidean distance between two matrices.

Let $\pi_k$ be a permutation of $\mathcal{S}$ such that
\begin{equation}
d_{\pi_k(1),k}
\le
d_{\pi_k(2),k}
\le
\cdots
\le
d_{\pi_k(|\mathcal{S}|),k}.
\end{equation}
Given a trimming hyperparameter $\mathtt{drop}$, where
$0 \le \mathtt{drop} < |\mathcal{S}|$, CIF keeps the
$r = |\mathcal{S}| - \mathtt{drop}$ languages closest to the centroid:
\begin{equation}
\mathcal{S}^{\mathrm{keep}}_k
=
\{\pi_k(j) \mid 1 \le j \le r\}.
\end{equation}
The fused residual intervention is the mean of the retained residuals:
\begin{equation}
\Delta \mathbf{W}_{\mathrm{CIF},k}
=
\frac{1}{|\mathcal{S}^{\mathrm{keep}}_k|}
\sum_{s \in \mathcal{S}^{\mathrm{keep}}_k}
\Delta \mathbf{W}_{s,k}.
\end{equation}
Finally, CIF constructs the intervention matrix as
\begin{equation}
\mathbf{W}_{\mathrm{CIF},k}(\alpha,\mathtt{drop})
=
\mathbf{I}_d
+
\alpha
\Delta \mathbf{W}_{\mathrm{CIF},k},
\end{equation}
where $\alpha \ge 0$ controls the intervention strength.
When $\alpha=0$, CIF reduces to the identity transformation.
When $\mathtt{drop}=0$, CIF uses the untrimmed mean of all language-specific residuals; if additionally $\alpha=1$, this is equivalent to averaging all language-specific projection matrices.

\subsection{Inference-Time Intervention}

During inference, let $\mathbf{h}_{q,k}$ be the hidden representation of input $q$ at layer $k$ and at the intervention position.
For sequence-level scoring, the model is given a complete input sequence and computes a score for the whole sequence, such as the log-likelihood of a candidate answer in multiple-choice evaluation.
In this case, we intervene on the hidden state of the last non-padding token, since this position summarizes the complete input sequence used for scoring.
For autoregressive generation, where tokens are generated one by one, we intervene on the hidden state at the current decoding position.
CIF applies the fused intervention as
\begin{equation}
\hat{\mathbf{h}}_{q,k}
=
\mathbf{W}_{\mathrm{CIF},k}(\alpha,\mathtt{drop})
\mathbf{h}_{q,k}.
\end{equation}
Equivalently, using the residual form,
\begin{equation}
\hat{\mathbf{h}}_{q,k}
=
\mathbf{h}_{q,k}
+
\alpha
\Delta \mathbf{W}_{\mathrm{CIF},k}
\mathbf{h}_{q,k}.
\end{equation}
The intervened representation $\hat{\mathbf{h}}_{q,k}$ replaces $\mathbf{h}_{q,k}$ before the forward computation.
For all layers, the same replacement is performed independently at each layer.
Because $\mathbf{W}_{\mathrm{CIF},k}$ is shared across languages, CIF does not require a language identifier or a separate projection matrix at inference time.

\begin{table*}[t!]
    \centering
    \tiny
    \begin{tabular}{l|cccccccc}
        \toprule
        \textbf{Method} &
        \textbf{XCOPA} & \textbf{XStoryCloze} & \textbf{XWinograd} & \textbf{XCSR} &
        \textbf{XNLI} & \textbf{MZSRE} & \textbf{WMT} & \textbf{Flores} \\
        \midrule

        \textbf{Zero-shot} 
        & 53.17 
        & \cellcolor{lightgray}\underline{81.39} 
        & 43.61 
        & 39.12 
        & 56.06 
        & 39.98 
        & 39.57 
        & 38.45 \\

        \textbf{Translation-NLLB} 
        & \cellcolor{lightblue}\textbf{59.41} 
        & 76.29 
        & \cellcolor{lightblue}\textbf{52.45} 
        & \cellcolor{lightblue}\textbf{44.02} 
        & 49.49 
        & \cellcolor{lightblue}\textbf{42.81} 
        & 23.60 
        & 32.81 \\

        \midrule
        \rowcolor{gray!20}
        \multicolumn{9}{l}{\textbf{Intervention Methods}} \\

        \textbf{ITI} 
        & 55.20\textsubscript{$\pm$0.28} 
        & 73.05\textsubscript{$\pm$0.66} 
        & 35.60\textsubscript{$\pm$1.03} 
        & 36.18\textsubscript{$\pm$0.14} 
        & 55.83\textsubscript{$\pm$0.31} 
        & 39.28\textsubscript{$\pm$0.22} 
        & 36.58\textsubscript{$\pm$0.57} 
        & 38.23\textsubscript{$\pm$0.26} \\

        \textbf{CAA} 
        & 55.24\textsubscript{$\pm$0.71} 
        & 71.58\textsubscript{$\pm$1.57} 
        & 38.74\textsubscript{$\pm$0.48} 
        & 31.10\textsubscript{$\pm$0.43} 
        & 53.57\textsubscript{$\pm$0.82} 
        & 38.45\textsubscript{$\pm$0.55} 
        & 37.44\textsubscript{$\pm$1.12} 
        & 37.54\textsubscript{$\pm$0.69} \\

        \textbf{SADI} 
        & 55.04\textsubscript{$\pm$0.39} 
        & 79.05\textsubscript{$\pm$0.88} 
        & 39.58\textsubscript{$\pm$0.74} 
        & 31.09\textsubscript{$\pm$0.19} 
        & 49.22\textsubscript{$\pm$0.63} 
        & 36.02\textsubscript{$\pm$0.27} 
        & 39.48\textsubscript{$\pm$0.59} 
        & 39.00\textsubscript{$\pm$0.36} \\

        \textbf{INCLINE} 
        & 53.98\textsubscript{$\pm$0.17} 
        & 81.15\textsubscript{$\pm$0.51} 
        & 46.85\textsubscript{$\pm$0.32} 
        & 38.60\textsubscript{$\pm$0.09} 
        & \cellcolor{lightgray}\underline{58.52\textsubscript{$\pm$0.06}} 
        & 40.62\textsubscript{$\pm$0.18} 
        & \cellcolor{lightgray}\underline{42.02\textsubscript{$\pm$0.33}} 
        & \cellcolor{lightgray}\underline{39.70\textsubscript{$\pm$0.11}} \\

        \multirow{2}{*}{\textbf{CIF}} 
        & \cellcolor{lightgray}\underline{57.36\textsubscript{$\pm$0.18}}* 
        & \cellcolor{lightblue}\textbf{82.76\textsubscript{$\pm$0.81}}* 
        & \cellcolor{lightgray}\underline{48.73\textsubscript{$\pm$0.16}}* 
        & \cellcolor{lightgray}\underline{40.19\textsubscript{$\pm$0.03}}* 
        & \cellcolor{lightblue}\textbf{60.32\textsubscript{$\pm$0.26}}* 
        & \cellcolor{lightgray}\underline{41.14\textsubscript{$\pm$0.08}}* 
        & \cellcolor{lightblue}\textbf{42.70\textsubscript{$\pm$0.46}}* 
        & \cellcolor{lightblue}\textbf{40.92\textsubscript{$\pm$0.13}}* \\

        & {\tiny (+3.38)}
        & {\tiny (+1.61)}
        & {\tiny (+1.88)}
        & {\tiny (+1.59)}
        & {\tiny (+1.80)}
        & {\tiny (+0.52)}
        & {\tiny (+0.68)}
        & {\tiny (+1.22)} \\

        \bottomrule
    \end{tabular}

    \caption{
    Cross-lingual reasoning and generation performance across multiple benchmarks.
    The best score in each column is shown in \cellcolor{lightblue}\textbf{bold blue},
    and the second-best score is shown in
    \cellcolor{lightgray}\underline{gray underlined}.
    Numbers in parentheses indicate the improvement of CIF over INCLINE.
    For intervention-based methods, we run three random seeds and report the mean and standard deviation. Statistical significance (*) is determined using a paired t-test with p=0.05.
    }

    \label{tab:crosslingual_results}
\end{table*}

\section{Experiments}

\subsection{Experimental Setup}

\paragraph{Benchmarks.}
We evaluate CIF on eight multilingual benchmarks that cover both understanding and generation.
For discriminative reasoning and sentence understanding, we use XCOPA~\citep{ponti-etal-2020-xcopa}, XStoryCloze~\citep{lin-etal-2022-shot}, XWinograd~\citep{tikhonov-ryabinin-2021-heads}, XCSR~\citep{lin-etal-2021-common}, and XNLI~\citep{conneau-etal-2018-xnli}.
For generation-oriented evaluation, we use MzsRE~\citep{wang-etal-2024-retrieval}, WMT23~\citep{kocmi-etal-2023-findings} and FLORES-101~\citep{goyal-etal-2022-flores}.
These benchmarks cover typologically diverse languages across multiple language families and scripts, including high-resource and low-resource languages.
Together, these benchmarks allow us to test whether CIF improves not only multilingual classification-style reasoning but also factual editing and machine translation.

\paragraph{Model Backbones.}
We conduct experiments on four multilingual LLM families: Qwen~\citep{bai2023qwen}, BLOOMZ~\citep{muennighoff-etal-2023-crosslingual}, LLaMA~\citep{touvron2023llama,touvron2023llama2,dubey2024llama3}, and Mistral~\citep{jiang2023mistral}.
For each backbone, we first learn language-specific projection matrices from parallel corpora, following Section~\ref{sec:3.1}.
INCLINE applies these matrices independently for each language, whereas CIF consolidates them into a shared language-shared intervention operator.
This comparison directly tests whether multilingual projection matrices contain transferable intervention representations that complement the language-specific alignment.

\paragraph{Baselines.}
We compare CIF with six baselines.
\textbf{Zero-shot} evaluates the original multilingual LLM without any intervention.
\textbf{Translation-NLLB} translates non-English inputs into English with NLLB~\citep{nllb-team-etal-2022-no}(nllb-200) before inference.
\textbf{ITI}~\citep{li2023inference} intervenes on attention heads associated with target behaviors.
\textbf{CAA}~\citep{rimsky-etal-2024-steering} builds fixed steering vectors from activation differences between positive and negative examples.
\textbf{SADI}~\citep{wang2025sadi} dynamically adjusts intervention vectors according to input semantics.
\textbf{INCLINE}~\citep{wang2025incline} learns language-specific projection matrices and applies them separately at inference time.
Implementation details are provided in Appendix~\ref{sec:more_implement}.

\begin{table*}[t!]
    \centering
    \tiny
    \begin{tabular}{lcc|cccccccc}
        \toprule
        \textbf{Method} &
        \textbf{Centroid} &
        \textbf{Drop} &
        \textbf{XCOPA} & \textbf{XStoryCloze} & \textbf{XWinograd} & \textbf{XCSR} &
        \textbf{XNLI} & \textbf{MZSRE} & \textbf{WMT} & \textbf{Flores} \\
        \midrule

        \textbf{CIF Variant}
        & \xmark & \xmark
        & 53.98
        & 81.15
        & 46.85
        & 38.60
        & 58.52
        & 40.62
        & \cellcolor{lightgray}\underline{42.02}
        & 39.70 \\

        \textbf{CIF Variant}
        &        & \xmark
        & \cellcolor{lightgray}\underline{55.81}
        & \cellcolor{lightblue}\textbf{83.11}
        & \cellcolor{lightgray}\underline{48.62}
        & \cellcolor{lightgray}\underline{40.00}
        & \cellcolor{lightgray}\underline{59.29}
        & \cellcolor{lightgray}\underline{41.08}
        & 41.86
        & \cellcolor{lightgray}\underline{40.08} \\

        \multirow{1}{*}{\textbf{CIF}}
        &        &
        & \cellcolor{lightblue}\textbf{57.36}
        & \cellcolor{lightgray}\underline{82.76}
        & \cellcolor{lightblue}\textbf{48.73}
        & \cellcolor{lightblue}\textbf{40.19}
        & \cellcolor{lightblue}\textbf{60.32}
        & \cellcolor{lightblue}\textbf{41.14}
        & \cellcolor{lightblue}\textbf{42.70}
        & \cellcolor{lightblue}\textbf{40.92} \\

        \bottomrule
    \end{tabular}

    \caption{
    Ablation study of CIF averaged over four backbone models.
    Cross marks indicate removed CIF representations.
    The best score in each column is shown in \cellcolor{lightblue}\textbf{bold blue},
    and the second-best score is shown in \cellcolor{lightgray}\underline{gray underlined}.
    Numbers in parentheses indicate the improvement of CIF over the variant with both centroid and drop removed.
    }

    \label{tab:cif_core_ablation}
\end{table*}

\subsection{Main Results}

Table~\ref{tab:crosslingual_results} presents the main results, where each entry reports the mean performance and variance across runs.
CIF achieves the highest mean overall performance among all intervention-based methods, reaching a macro-average score of $51.77$\% across benchmarks.
Compared with INCLINE, the strongest cross-lingual intervention baseline, CIF improves the overall average by up to 
\textbf{+1.73} 
percentage points~(pp).
It outperforms SADI, ITI, and CAA by \textbf{+5.71}, \textbf{+5.53}, and \textbf{+6.31} pp, respectively.
These gains show that fusing language-specific projection matrices into a shared operator is more effective than applying separate pairwise intervention operators.

\paragraph{Reasoning and Understanding.}
CIF consistently improves reasoning and sentence understanding.
Compared with INCLINE, CIF improves XCOPA by \textbf{+3.38} pp, XStoryCloze by \textbf{+1.61} pp, XWinograd by \textbf{+1.88} pp, XCSR by \textbf{+1.59} pp, and XNLI by \textbf{+1.80} pp.
The greatest gain appears in XCOPA, where the CIF increases the score from $53.98$\% to $57.36$\%.
These results suggest that shared cross-lingual intervention representations provide stable reasoning signals.

\paragraph{Factual Editing and Generation.}
CIF also improves generation-oriented tasks.
Compared with INCLINE, CIF improves MzsRE by \textbf{+0.52} pp, WMT23 by \textbf{+0.68} pp, and FLORES-101 by \textbf{+1.22} pp.
On both WMT23 and FLORES-101, CIF obtains the best performance among intervention-based methods.
These results indicate that the fused operator is not limited to classification-style tasks; it also benefits tasks that require faithful multilingual generation.

\paragraph{Comparison with Non-Intervention Baselines.}
Translation-NLLB is competitive on some reasoning benchmarks, such as XCOPA and XCSR, because translating inputs into English can reduce part of the cross-lingual gap.
Translation-NLLB is a strong practical pipeline when high-quality translation is available. Our CIF method provides a translation-free and scalable alternative for model-internal cross-lingual intervention. Translation-NLLB is preferable when an error-free translation system is available for the task’s target language that correctly translates the meaning of the language. In other cases our CIF method is a viable alternative that directly operates in the model representation space.

For generation-oriented benchmarks such as WMT23 and FLORES-101, an external translation pipeline may introduce translation artifacts or change the original generation objective.
In contrast, CIF intervenes directly in the model representation space and does not rely on an external translation system.
The results support our hypothesis that language-specific projection matrices share transferable representations that can be unified for cross-lingual intervention.

\subsection{Ablation Study}

Table~\ref{tab:cif_core_ablation} presents the ablation results averaged across four backbone models.
We examine two key representations of CIF: centroid fusion and trimmed aggregation.

\paragraph{Effect of Centroid Fusion.}
Without centroid fusion, the method keeps language-specific projection matrices independent, following the INCLINE-style intervention strategy.
This variant obtains the weakest overall performance.
Adding centroid fusion improves most benchmarks, increasing XCOPA from $53.98$\% to $55.81$\% by \textbf{+1.83} pp, XStoryCloze by \textbf{+1.96} pp, and XNLI by \textbf{+0.77} pp.
These improvements show that projection matrices learned from different languages share common representations, and that explicitly consolidating these improves cross-lingual generalization.

\paragraph{Effect of Trimmed Aggregation.}
The full-fledged
CIF model further applies trimmed aggregation, which removes projection matrices farthest from the centroid before constructing the final intervention operator.
This step further improves XCOPA by \textbf{+1.55} pp, XNLI by \textbf{+1.03} pp, and both WMT23 and FLORES-101 by \textbf{+0.84} BLEU points.
These gains suggest that some outlier projections may be noisy or conflict with the shared multilingual representations.
By filtering these unstable projections, trimmed aggregation makes the fused operator more robust.

\section{Analysis}
\label{sec:analysis}

CIF assumes that language-specific interventions share cross-lingual signals, while still allowing language-dependent deviations that may help or hurt fusion.
This assumption is consistent with prior work showing that multilingual models encode both language-neutral and language-specific structure \citep{pires-etal-2019-multilingual,libovicky-etal-2020-language,zhao-etal-2021-inducing}.
We analyze whether CIF follows this behavior from four perspectives: projection-space geometry, centroid-guided dropping, representation-level alignment, and error analysis.
We put more analysis including sensitivity to $\alpha$ and $\mathtt{drop}$ in Appendix~\ref{sec:alpha_drop_sensitivity}, language-family effects in Appendix~\ref{sec:family_analysis}, leave-one-language-out transfer in Appendix~\ref{sec:leave_one_language}, per-language effects in Appendix~\ref{sec:per_language}, and case study in Appendix~\ref{sec:case_study}.

\begin{figure}[htbp]
\centering
\subfloat[\scriptsize{Centroid alignment.}]{
\includegraphics[width=0.47\textwidth]{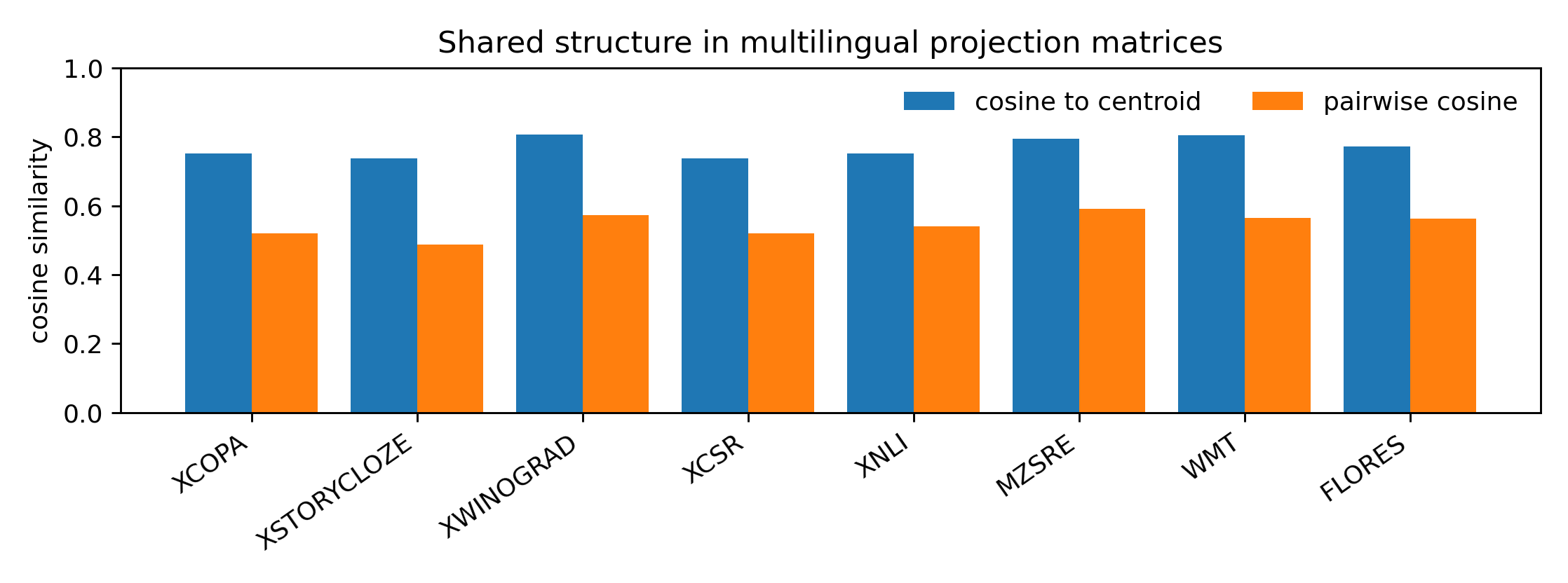}
\label{fig:projection_shared_structure}
}
\hfil
\subfloat[\scriptsize{Spectral concentration.}]{
\includegraphics[width=0.47\textwidth]{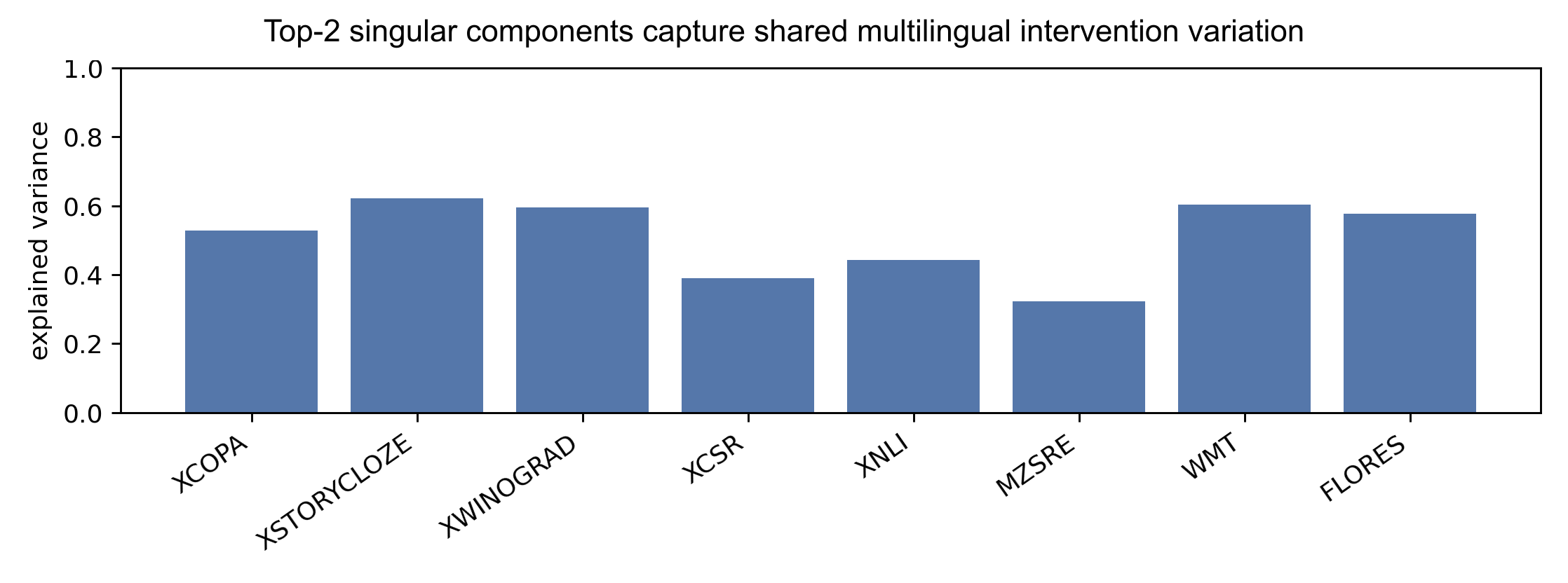}
\label{fig:projection_svd}
}
\caption{
Projection-space evidence for shared cross-lingual intervention 
representations.
Language-specific projection perturbations are consistently aligned with the centroid, and a small number of singular vectors explain a substantial amount of projection variance.
}
\label{fig:cif_shared_projection_structure}
\end{figure}

\subsection{Shared Structure in Projection Space}
\label{sec:projection_space_analysis}

We first ask whether projections learned from different languages contain a shared intervention representation.
Because each projection matrix includes an identity mapping, we remove this trivial part and analyze only the learned residual update.
We then flatten each residual projection into a vector and study its multilingual geometry.

\paragraph{Centroid alignment.}

We begin by testing whether language-specific residuals cluster around a common center.
For each dataset, we compute the mean residual vector across languages and compare two quantities: the average similarity between each language and this centroid, and the average similarity between pairs of languages.

Figure~\ref{fig:projection_shared_structure} shows that language-specific residuals are consistently more aligned with the centroid than with one another.
This result should be interpreted as a descriptive diagnostic, as the centroid is estimated from the same set of languages.
Even so, the representation suggests that the learned projections are not scattered independently.
They share a stable central representation, and each language adds its own deviation around it.
This supports the main design of CIF: fusion should preserve the shared multilingual representation without forcing all languages to use exactly the same projection.

\paragraph{Spectral structure.}

Next, we examine whether this shared structure is compact.
To do so, we stack the residual vectors from all languages into a language-by-parameter matrix and measure how much projection energy is captured by the leading singular vectors after singular value decomposition.

As shown in Figure~\ref{fig:projection_svd}, a small number of representations
explain a large fraction of the projection energy, especially on FLORES, XStoryCloze, WMT, and XWinograd \citep{goyal-etal-2022-flores,lin-etal-2022-shot,kocmi-etal-2024-findings,tikhonov-ryabinin-2021-heads}.
This indicates that multilingual projections are not arbitrary language-specific transformations.
Instead, they concentrate in a low-dimensional shared subspace.
CIF is therefore not simply averaging unrelated matrices; it aims to keep the dominant cross-lingual signals, while reducing outlier projections.

\begin{figure}[htbp]
\centering
\includegraphics[width=0.5\textwidth]{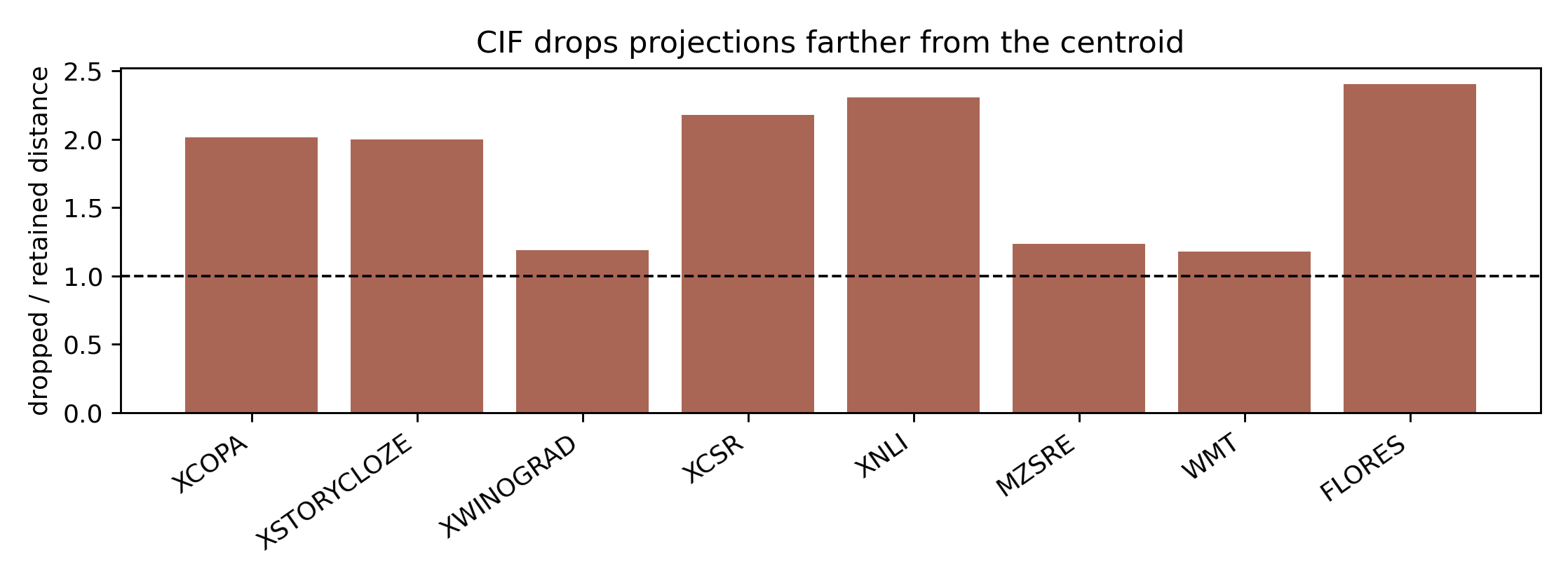}
\caption{
Centroid-guided outlier filtering in CIF.
Projection matrices removed by CIF are consistently farther from the centroid than retained matrices, indicating that the drop operation filters outlier deviations rather than discarding projections arbitrarily.
}
\label{fig:cif_centroid_drop_ratio}
\end{figure}

\subsection{Does Centroid-Guided Drop Remove Outliers?}
\label{sec:centroid_drop_analysis}

Averaging all projections assumes that each language contributes an equally compatible intervention.
This assumption can fail when some residuals lie far from the shared linguistic structure and introduce conflicting representations.
CIF addresses this by dropping the residuals that are farthest from the centroid before fusion.

We test whether this step behaves as intended by comparing the centroid distance of dropped and retained projections.
Figure~\ref{fig:cif_centroid_drop_ratio} shows that the dropped projections have consistently larger centroid distances than the retained projections in all datasets.
This difference 
is especially clear in FLORES, XNLI, XCSR, and XCOPA datasets \citep{goyal-etal-2022-flores,conneau-etal-2018-xnli,lin-etal-2021-common,ponti-etal-2020-xcopa}.
This suggests that the centroid-guided drop acts as a geometric outlier filter rather than an arbitrary language removal strategy.

We further examine which languages are removed most often.
The removed languages vary across datasets: Tamil is most frequently removed on XCOPA, Hindi on XStoryCloze, XCSR, XNLI, and FLORES, and Japanese on WMT.
This dataset-specific representation suggests that CIF does not depend on a fixed language list.
Instead, the drop decision adapts to the task-specific geometry of the learned projections.
A language may be compatible with the shared intervention structure in one task, but become an outlier in another.

\begin{figure}[htbp]
\centering
\subfloat[\scriptsize{Before intervention.}]{
\includegraphics[width=0.22\textwidth]{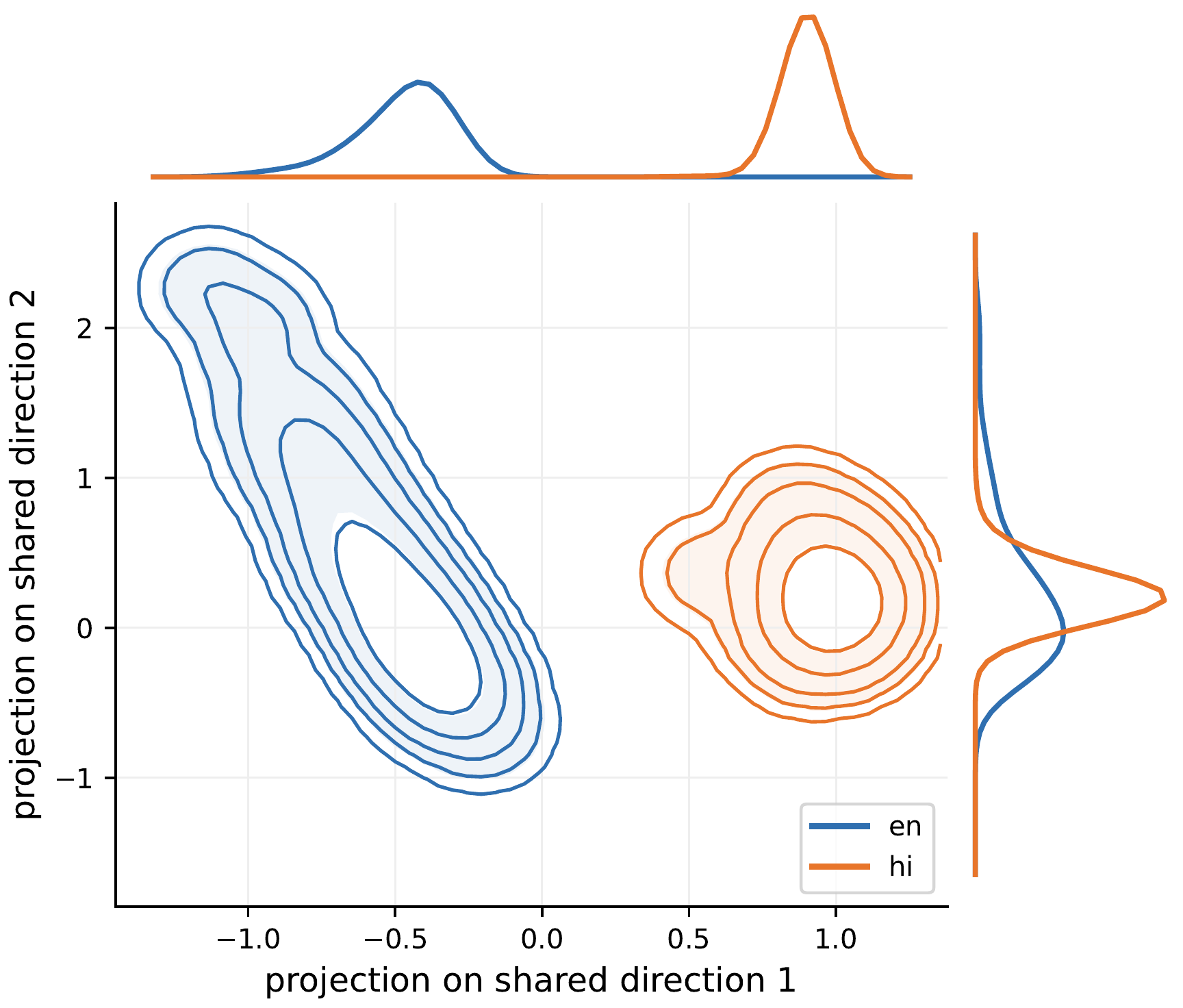}
\label{fig:cif_activation_before}
}
\hfil
\subfloat[\scriptsize{After CIF intervention.}]{
\includegraphics[width=0.22\textwidth]{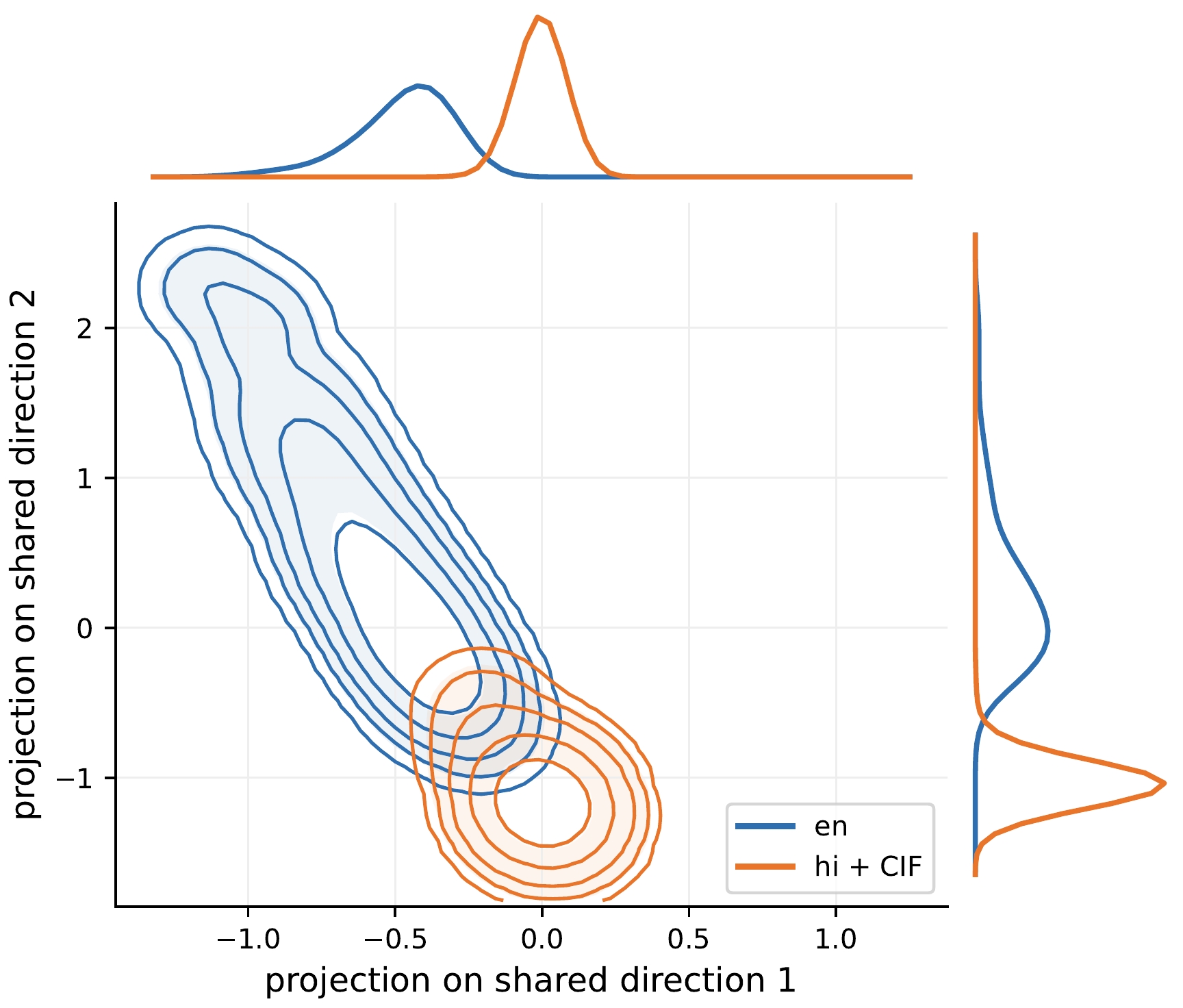}
\label{fig:cif_activation_after}
}
\caption{
Representation-level effect of CIF on English-Hindi FLORES examples.
Before intervention, the hidden representations form separated distributions.
After applying CIF, the Hindi distribution moves closer to the English distribution and even overlaps with it.
}
\label{fig:cif_activation_kde}
\end{figure}

\begin{table}[t]
    \centering
    \small
    \begin{tabular}{l|c}
        \toprule
        \textbf{Error Type} & \textbf{\# Examples} \\
        \midrule
        CIF fixes INCLINE & 66 \\
        CIF fixes CIF w/o drop & 59 \\
        CIF fixes both INCLINE and CIF w/o drop & 3 \\
        CIF regresses vs. INCLINE & 41 \\
        CIF regresses vs. CIF w/o drop & 38 \\
        \bottomrule
    \end{tabular}
    \caption{
    Error analysis of the randomly sampled XCOPA subset.
    }
    \label{tab:cif_error_transitions}
\end{table}

\subsection{Representation-Level Alignment}
\label{sec:representation_alignment}

The projection-space results suggest that the CIF extracts the structure shared across languages.
We next examine whether this shared structure is also reflected in the model's hidden representations.
Specifically, we test whether applying CIF makes parallel English and Hindi representations closer at the intervention layer.

We perform this analysis with FLORES with Qwen2.5-7B-Instruct~\citep{yang2025qwen25}.
We use English--Hindi sentence pairs because Hindi is repeatedly identified as an outlying language in the centroid-drop analysis.
For 260 parallel sentence pairs, we extract the MLP hidden state at the intervention layer before and after applying CIF.
We then project all representations into two dimensions using PCA for visualization.

Figure~\ref{fig:cif_activation_kde} shows the representation distributions.
Before CIF, English and Hindi form clearly separated regions, indicating strong language-specific variation in the hidden space.
After CIF, the Hindi representations shift toward the English representations.
This shows that CIF changes not only the final predictions, but also the geometry of the internal representation space.

Distance metrics show the same trend.
After CIF, the English--Hindi centroid Euclidean distance decreases from 5.45 to 2.53, while the average cross-language pairwise distance decreases from 5.50 to 2.54.
Both distances are reduced by about 54\%.
These results support the projection-space finding: CIF suppresses outlier variation while preserving shared cross-lingual signals.

\subsection{Error Analysis}
\label{sec:error_analysis}

We further examine how CIF changes model behavior at the instance level.
We perform an error analysis using the XCOPA data and the Qwen2.5-7B-Instruct backbone model.
We randomly sample 40 examples from each of the 9 languages, resulting in 360 examples.
All baseline methods and our CIF method are evaluated by scoring the two candidate answer choices and selecting the higher-scoring option.
We compare CIF without dropping (i.e., projection fusion without the centroid-guided drop step), INCLINE, and full-fletched CIF.

Table~\ref{tab:cif_error_transitions} shows the corresponding prediction transitions.
CIF corrects 66 examples that INCLINE predicts incorrectly and 59 examples that CIF without dropping predicts incorrectly.
Among them, 3 examples are wrongly predicted by both baselines but corrected by CIF.
Although CIF introduces some new errors, it corrects substantially more than it adds.
This imbalance suggests that centroid-guided fusion changes the intervention representations in a beneficial way, rather than merely increasing the intervention strength.


\section{Conclusion}
We presented \textbf{Centroid Intervention Fusion (CIF)}, an inference-time framework for cross-lingual intervention.
CIF constructs a shared cross-lingual intervention operator by fusing projection matrices learned from multiple language pairs.
Instead of maintaining separate operators for different language pairs, CIF estimates a robust multilingual centroid in the perturbation space and uses centroid-guided trimming to reduce the effect of outlying or conflicting projections.
Across multilingual reasoning, natural language inference, factual editing, and machine translation tasks, CIF improves over previous intervention methods across multiple LLM families.
Further analysis shows that projection matrices from different languages contain stable shared structures, and that CIF can use these structures to produce better transferable intervention representations.
These results suggest that cross-lingual intervention can be modeled in a shared projection space, providing a scalable alternative to pairwise intervention while preserving the original multilingual representations of the model.

\section{Limitations}
Although CIF shows consistent gains across tasks and model families, several limitations remain.
First, our experiments focus on text-based multilingual understanding and generation; whether CIF transfers to multimodal or interactive settings remains unclear.
Second, CIF uses a fixed intervention strength and a fixed centroid-based filtering strategy within each task setting.
Adaptive strategies that account for language- or task-specific variation may further improve robustness.

\section*{Acknowledgements}
This research was funded by the AIDAVA (EU HORIZON-HLTH-2021-TOOL-06-03) and the CHIST-ERA iTRUST project(CHIST-ERA Call 2021, FWO grant G0L0822N).

The computations described in this research were performed using the The Flemish Supercomputer Center Tier-1 HPC service~(https://www.vscentrum.be/) and the high-performance computing resources provided by CINECA~(https://www.cineca.it/), Italy.

\bibliography{custom}

\appendix

\section{Appendix}
\label{sec:appendix}

\subsection{Case Study}
\label{sec:case_study}

Table~\ref{tab:cif_case_study} compares CIF without dropping, INCLINE, and CIF on three XCOPA examples.
English translations are added only for readability.

The examples show two complementary effects.
In the Estonian and Indonesian cases, projection fusion already gives the correct prediction:
a spam-filled inbox supports deleting spam, and an emptied pocket supports taking out a ticket stub.
INCLINE instead selects the distractor, while CIF preserves the correct fusion signal.

The Swahili case shows why dropping is still useful.
CIF without dropping selects the wrong continuation, whereas INCLINE predicts correctly.
After centroid-guided dropping, CIF removes conflicting multilingual perturbations and recovers the correct answer.
These cases suggest that CIF can retain useful projected signals when they are reliable, and trim them when they become misleading.

\begin{table*}[t]
    \centering
    \scriptsize
    \begin{tabular}{l|p{2.7cm}|p{2.7cm}|p{2.7cm}|c|c|c|c}
        \toprule
        \textbf{Lang.} &
        \textbf{Premise} &
        \textbf{Choice A} &
        \textbf{Choice B} &
        \textbf{Gold} &
        \textbf{CIF w/o drop} &
        \textbf{INCLINE} &
        \textbf{CIF} \\
        \midrule

        et &
        Mehe e-postkast oli r\"ampsposti t\"ais. \newline
        \textit{(The man's email inbox was full of spam.)} &
        Ta kustutas r\"ampsposti. \newline
        \textit{(He deleted the spam.)} &
        Ta saatis laiali massimeili. \newline
        \textit{(He sent out a mass email.)} &
        A & A & B & \textbf{A} \\
        \hline
        id &
        Saya kosongkan kantong saya. \newline
        \textit{(I emptied my pocket.)} &
        Saya ambil sebuah potongan tiket. \newline
        \textit{(I took out a ticket stub.)} &
        Saya temukan sebuah senjata. \newline
        \textit{(I found a weapon.)} &
        A & A & B & \textbf{A} \\
        \hline
        sw &
        Msichana alitaka kumshukuru mwalimu wake wa hesabu. \newline
        \textit{(The girl wanted to thank her math teacher.)} &
        Msichana alibakia kizuizini baada ya shule. \newline
        \textit{(The girl stayed after school for detention.)} &
        Msichana alimletea tofaa mwalimu wake. \newline
        \textit{(The girl brought her teacher an apple.)} &
        B & A & B & \textbf{B} \\

        \bottomrule
    \end{tabular}
    \caption{
XCOPA case study.
CIF preserves correct unfiltered-fusion predictions in Estonian and Indonesian, while correcting an unfiltered-fusion error in Swahili.
English translations are added for readability.
}
    \label{tab:cif_case_study}
\end{table*}

\begin{table*}[t]
    \centering
    \small
    \begin{tabular}{l|l|p{8.2cm}|c|c}
        \toprule
        \textbf{Dataset} & \textbf{Output} & \textbf{Prompt} & \textbf{Metric} & $\mathbf{|L|}$ \\
        \midrule

        XCOPA &
        2-way class &
        Here is a premise: ``\{premise\}''. A: ``\{choice1\}'' B: ``\{choice2\}'' 
        What is the \{question\}? ``A'' or ``B''? &
        F1 & 10 \\

        \midrule
        XStoryCloze &
        2-way class &
        \{input\} What is a possible continuation for the story given the following options?
        A: \{ending1\} B: \{ending2\} &
        F1 & 8 \\

        \midrule
        XWinograd &
        2-way class &
        \{input\} Replace the blank in the above sentence with the correct option:
        A: \{option1\} B: \{option2\} &
        F1 & 6 \\

        \midrule
        XCSR &
        multi-choice &
        Question: \{question\} \{choices\} Answer: &
        F1 & 15 \\

        \midrule
        XNLI &
        3-way class &
        Take the following as truth: \{premise\}
        Then the following statement: ``\{hypothesis\}'' is ``entailment'', ``neutral'', or ``contradiction''? &
        F1 & 14 \\

        \midrule
        MZSRE &
        answer &
        \{context\} Question: \{question\} Answer: &
        F1 & 10 \\

        \midrule
        WMT23 &
        answer &
        Translate the following sentence from \{language\} to English: \{input\} &
        BLEU & 6 \\

        \midrule
        FLORES-101 &
        answer &
        Translate the following sentence from \{language\} to English: \{input\} &
        BLEU & 10 \\

        \bottomrule
    \end{tabular}
    \caption{
    Downstream evaluation tasks used in our cross-lingual intervention experiments.
    $|L|$ denotes the number of evaluated languages.
    We report F1 for classification and factual editing tasks, and BLEU for machine translation tasks.
    }
    \label{tab:downstream_tasks}
\end{table*}

\begin{table*}[t!]
    \centering
    \scriptsize
    \begin{tabular}{lllccl}
        \toprule
        \textbf{Dataset} & \textbf{Language} & \textbf{Group} &
        \textbf{INCLINE} & \textbf{CIF} & \textbf{Gain} \\
        \midrule

        XCSR & hi & low-resource / distant / trimmed
        & 29.98 & 42.28 & \textbf{+12.30 pp} \\
        XCSR & ar & distant
        & 37.78 & 49.22 & \textbf{+11.45 pp} \\
        XCSR & vi & low-resource
        & 39.58 & 51.53 & \textbf{+11.95 pp} \\
        XCSR & zh & distant
        & 45.70 & 56.58 & \textbf{+10.88 pp} \\
        XCSR & sw & low-resource
        & 35.50 & 39.37 & \textbf{+3.87 pp} \\

        \midrule
        XNLI & tr & distant
        & 56.64 & 63.56 & \textbf{+6.93 pp} \\
        XNLI & ar & distant
        & 59.19 & 65.55 & \textbf{+6.37 pp} \\
        XNLI & sw & low-resource
        & 50.62 & 56.74 & \textbf{+6.12 pp} \\
        XNLI & hi & low-resource / distant / trimmed
        & 56.42 & 61.78 & \textbf{+5.36 pp} \\
        XNLI & th & low-resource / distant
        & 55.26 & 60.50 & \textbf{+5.24 pp} \\

        \midrule
        XStoryCloze & id & low-resource
        & 78.13 & 82.10 & \textbf{+3.97 pp} \\
        XStoryCloze & zh & distant
        & 78.11 & 81.95 & \textbf{+3.84 pp} \\
        XStoryCloze & hi & low-resource / distant / trimmed
        & 82.05 & 83.01 & \textbf{+0.96 pp} \\

        \midrule
        FLORES & ar & distant
        & 36.07 & 41.33 & \textbf{+5.25 BLEU} \\
        FLORES & hi & low-resource / distant / trimmed
        & 47.00 & 50.99 & \textbf{+3.99 BLEU} \\
        WMT23 & ja & distant
        & 44.49 & 45.27 & \textbf{+0.79 BLEU} \\
        WMT23 & zh & distant
        & 45.91 & 46.13 & \textbf{+0.22 BLEU} \\

        \midrule
        XCOPA & ta & low-resource / distant / trimmed
        & 56.02 & 56.95 & \textbf{+0.93 pp} \\
        XCOPA & th & low-resource / distant
        & 53.05 & 53.50 & \textbf{+0.45 pp} \\
        XCOPA & zh & distant
        & 54.95 & 53.25 & $-1.70$ pp \\
        FLORES & zh & distant
        & 52.24 & 49.03 & $-3.21$ pp \\
        WMT23 & ru & distant
        & 60.86 & 57.18 & $-3.68$ pp \\

        \bottomrule
    \end{tabular}

    \caption{
    Per-language comparison between CIF and INCLINE.
    Scores are averaged over the four backbones after selecting a single intervention setting for each dataset--backbone pair.
    F1-based tasks are reported in percentages, while WMT23 and FLORES are reported in BLEU.
    The table includes low-resource, typologically distant, and frequently trimmed languages to examine whether CIF benefits languages beyond those retained in the fused operator.
    }
    \label{tab:per_language_cif}
\end{table*}

\subsection{Implementation Details}
\label{sec:more_implement}
\paragraph{Experimental Setup}Following prior cross-lingual intervention settings~\citep{wang2025incline}, we learn language-specific projection matrices from English--non-English parallel sentences.
For each English--target-language pair, we randomly sample 500 sentence pairs.
We use News Commentary v16~\citep{barrault-etal-2019-findings} when the language pair is available, and use CCAligned~\citep{el-kishky-etal-2020-ccaligned} for languages not covered by News Commentary.
For each sentence pair, we extract the representation of the last token from every transformer layer.
For each layer $l$, we solve an ordinary least-squares problem to learn a projection matrix $W_l$ that maps the non-English representation to the corresponding English representation.
We do not update the backbone parameters; all projection matrices are learned offline and applied only at inference time.

At inference time, we intervene once at the last input token and apply the intervention across all transformer layers.
Following prior work, we tune the intervention strength $\alpha$ over a fixed non-zero grid from $-1$ to $1$ with step of $0.1$:
For each dataset--backbone pair, we select a single $\alpha$ according to average validation performance across all target languages on the validation set of each downstream task.
The selected $\alpha$ is then shared by all languages in that setting, preventing language-specific hyperparameter tuning.
For CIF, we additionally tune the number of removed projection matrices over a small grid using the validation set of the downstream task, $\mathrm{drop}\in\{0,1,2\}$, where $\mathrm{drop}=0$ corresponds to CIF without trimming.
The same drop value is selected at the dataset--backbone level and shared across all target languages for a specific downstream task.

Following the prior work~\cite{wang2025incline}, we use the prompt templates listed in Table~\ref{tab:downstream_tasks} for all methods.
For classification-style tasks, including XCOPA, XStoryCloze, XWinograd, XCSR, and XNLI, we score the fixed answer candidates and select the candidate with the highest model score.
For generation-style tasks, including MZSRE, WMT23, and FLORES-101, we use greedy decoding with temperature $0$.
For machine translation tasks, we evaluate generated translations using SacreBLEU~\citep{post-2018-call}.
For the translation baseline, we use NLLB-200 distilled 600M (\texttt{facebook/nllb-200-distilled-600M}) to translate non-English inputs into English before evaluating them with the same downstream backbone.

All experiments are conducted on NVIDIA-A100 (80GB).

\paragraph{Baseline Setup}
For all intervention baselines, we use the same target languages, backbone models, and tuning budget as CIF.
Specifically, ITI, CAA, SADI, INCLINE, and CIF all tune the intervention strength $\alpha$ on the same non-zero grid and select one shared $\alpha$ for each dataset--backbone pair according to average validation performance across target languages.
No method is allowed to choose language-specific intervention strengths.
For methods with no additional hyperparameters, such as ITI, CAA, and SADI in our implementation, this gives the same tuning budget as INCLINE.

We adapt ITI, CAA, and SADI to the multilingual setting using the same inference-time intervention interface.
ITI is implemented as a representation-based activation intervention, where the steering representation is estimated from the same English--non-English parallel representations used to construct the projection matrices.
CAA is implemented as contrastive activation steering over the corresponding English and non-English representations.
SADI is implemented as a representation-level steering baseline that applies the learned source--target activation representation at inference time.
For a fair comparison, all baselines intervene at the same layer positions and token position as CIF, use the same prompt templates and decoding settings, and are evaluated on the same target-language examples.
Backbone parameters are frozen for all methods.
This ensures that performance differences come from the intervention operator rather than from different training data, validation data, language-specific tuning, or decoding choices.

\subsection{Per-language Effects}
\label{sec:per_language}

We further analyze CIF at the language level using the selected intervention setting for each dataset--backbone pair.
This evaluation includes both retained and trimmed languages; a language whose projection matrix is removed during CIF construction is still evaluated at test time.

Table~\ref{tab:per_language_cif} reports representative low-resource, typologically distant, and frequently trimmed languages.
CIF yields particularly strong improvements on several challenging languages, including Arabic, Hindi, Vietnamese, and Chinese on XCSR, as well as Thai, Turkish, Swahili, and Hindi on XNLI.
For example, CIF improves Hindi on XCSR by 12.30 points, Arabic on XCSR by 11.45 points, and Arabic on FLORES by 5.25 BLEU.
These results suggest that the fused centroid captures transferable cross-lingual intervention representations that generalize beyond the languages most central to the fusion process.

However, the gains are not uniform across all target languages.
For instance, Chinese on FLORES decreases by 3.21 BLEU.
This indicates that trimming can sometimes remove language-specific information that is beneficial for individual languages.
Overall, the results suggest that CIF primarily improves the robustness of the shared cross-lingual intervention representation, rather than guaranteeing consistent gains for every language.

\subsection{Leave-one-language-out Transfer.}
\label{sec:leave_one_language}
We further evaluate whether CIF transfers to languages that do not participate in constructing the fused operator.
For each held-out language, we remove its projection matrix before fusion and evaluate the resulting operator on that language.
As shown in Table~\ref{tab:loo_cif}, LOO-CIF remains close to full CIF on most held-out languages.
On XCOPA, removing Tamil and Vietnamese even slightly improves the held-out scores by $+1.40$ and $+0.60$ points, respectively.
On FLORES, removing Hindi and Vietnamese even slightly improves the held-out scores by $+0.63$ and $+0.47$ points, respectively.
These results suggest that CIF captures intervention representations that transfer across languages rather than relying only on target-language-specific projections.
However, Chinese on FLORES drops by $-0.75$ BLEU, indicating that typologically distant languages may still benefit from retaining their own projection information.

\begin{table}[t!]
    \centering
    \scriptsize
    \begin{tabular}{llccc}
        \toprule
        \textbf{Dataset} & \textbf{Held-out Lang.} &
        \textbf{Full CIF} & \textbf{LOO-CIF} & \textbf{Gap} \\
        \midrule
        XCOPA  & ta & 52.80 & 54.20 & $+1.40$ \\
        XCOPA  & vi & 61.40 & 62.00 & $+0.60$ \\
        XCOPA  & zh & 70.80 & 70.40 & $-0.40$ \\
        \midrule
        FLORES & hi & 55.77 & 55.14 & $+0.63$ \\
        FLORES & vi & 45.33 & 44.83 & $+0.47$ \\
        FLORES & zh & 38.54 & 37.79 & $-0.75$ \\
        \bottomrule
    \end{tabular}
    \caption{
    Leave-one-language-out evaluation of CIF.
    The held-out language is excluded when constructing the fused operator but is still used as the evaluation target.
    XCOPA scores are F1 percentages and FLORES scores are BLEU scores.
    }
    \label{tab:loo_cif}
\end{table}

\begin{table*}[t!]
    \centering
    \scriptsize
    \begin{tabular}{lllc}
        \toprule
        \textbf{Dataset} & \textbf{Language} & \textbf{Linguistic Property} & \textbf{CIF Gain over INCLINE} \\
        \midrule

        XCSR & ar & Semitic / non-Latin script & $+11.45$ pp \\
        XCSR & hi & Indo-Aryan / non-Latin script & $+12.30$ pp \\
        XCSR & ja & Japonic / non-Latin script & $+9.58$ pp \\
        XCSR & ur & Indo-Aryan / non-Latin script & $+7.65$ pp \\
        XCSR & vi & Austroasiatic / low-resource setting & $+11.95$ pp \\
        XCSR & zh & Sino-Tibetan / non-Latin script & $+10.88$ pp \\

        \midrule
        XNLI & ar & Semitic / non-Latin script & $+6.37$ pp \\
        XNLI & hi & Indo-Aryan / non-Latin script & $+5.36$ pp \\
        XNLI & sw & Niger-Congo / low-resource setting & $+6.12$ pp \\
        XNLI & th & Kra-Dai / non-Latin script & $+5.24$ pp \\
        XNLI & tr & Turkic & $+6.93$ pp \\
        XNLI & ur & Indo-Aryan / non-Latin script & $+3.35$ pp \\
        XNLI & vi & Austroasiatic / low-resource setting & $+5.37$ pp \\
        XNLI & zh & Sino-Tibetan / non-Latin script & $+4.72$ pp \\

        \midrule
        FLORES & ar & Semitic / non-Latin script & $+5.25$ BLEU \\
        FLORES & hi & Indo-Aryan / non-Latin script & $+3.99$ BLEU \\
        FLORES & zh & Sino-Tibetan / non-Latin script & $-3.21$ BLEU \\

        \bottomrule
    \end{tabular}
    \caption{
    Representative per-language gains of CIF over INCLINE for typologically distant, non-Latin-script, or lower-resource languages.
    Classification tasks are reported in percentage points, while FLORES is reported in BLEU.
    CIF improves many distant languages, suggesting that the fused operator captures transferable intervention representations beyond closely related language families.
    However, the regression on Chinese in FLORES shows that language-specific information can still be important for open-ended generation.
    }
    \label{tab:family_analysis}
\end{table*}

\begin{table}[t!]
    \centering
    \scriptsize
    \begin{tabular}{lcccc}
        \toprule
        \textbf{Trimming Size} & \textbf{XCOPA} & \textbf{XNLI} & \textbf{WMT23} & \textbf{FLORES} \\
        \midrule
        $\mathrm{drop}=0$ & 54.81 & 59.29 & 41.86 & 40.08 \\
        $\mathrm{drop}=1$ & 56.98 & 64.03 & 42.45 & 40.69 \\
        $\mathrm{drop}=2$ & 56.84 & 63.39 & 42.24 & 40.90 \\
        \bottomrule
    \end{tabular}
    \caption{
    Sensitivity of CIF to the trimming size $\mathrm{drop}$.
    Scores are averaged over four backbones.
    XCOPA and XNLI are reported in F1 percentage, while WMT23 and FLORES are reported in BLEU.
    }
    \label{tab:drop_sensitivity}
\end{table}

\begin{table*}[t!]
    \centering
    \scriptsize
    \begin{tabular}{lcccccccc}
        \toprule
        \textbf{Dataset} &
        $\boldsymbol{\alpha=-1.0}$ &
        $\boldsymbol{\alpha=-0.7}$ &
        $\boldsymbol{\alpha=-0.4}$ &
        $\boldsymbol{\alpha=-0.1}$ &
        $\boldsymbol{\alpha=0.1}$ &
        $\boldsymbol{\alpha=0.4}$ &
        $\boldsymbol{\alpha=0.7}$ &
        $\boldsymbol{\alpha=1.0}$ \\
        \midrule
        XCOPA  & 55.54 & 55.44 & 55.62 & 55.17 & 55.14 & 55.03 & 54.88 & 54.53 \\
        XNLI   & 52.62 & 52.41 & 53.01 & 57.72 & 56.66 & 53.38 & 53.23 & 53.15 \\
        WMT23  & 32.44 & 34.52 & 33.65 & 40.46 & 40.97 & 42.24 & 39.48 & 38.62 \\
        FLORES & 28.46 & 32.56 & 37.79 & 38.57 & 38.00 & 37.57 & 37.64 & 37.90 \\
        \bottomrule
    \end{tabular}
    \caption{
    Sensitivity of CIF to intervention strength $\alpha$.
    Each entry reports the average score under a fixed $\alpha$ value across four backbones.
    Classification tasks are reported in F1 percentage, while WMT23 and FLORES are reported in BLEU.
    }
    \label{tab:alpha_sensitivity}
\end{table*}

\subsection{Effect of Language Family and Typological Distance}
\label{sec:family_analysis}

A natural question is whether CIF mainly benefits languages that are closely related to the source languages used to construct the fused operator, or whether it can also transfer to typologically distant languages.
To investigate this, we analyze per-language gains over INCLINE and group representative languages by broad linguistic properties.
We focus on languages that are either typologically distant from English, written in non-Latin scripts, or commonly considered lower-resource in multilingual benchmarks.
The goal of this analysis is not to define a strict typological taxonomy, but to examine whether CIF only helps languages close to English or whether its gains extend beyond related language families.

Table~\ref{tab:family_analysis} shows that CIF improves many typologically distant languages.
On XCSR, CIF obtains large gains for Arabic, Hindi, Japanese, Vietnamese, Urdu, and Chinese, with improvements ranging from $+7.65$ to $+12.30$ points.
On XNLI, CIF also improves distant or lower-resource languages such as Arabic, Hindi, Swahili, Thai, Turkish, Urdu, Vietnamese, and Chinese.
These results suggest that CIF does not merely exploit similarities among closely related Indo-European languages.
Instead, the fused operator appears to capture intervention representations that transfer across language families and scripts.

At the same time, typological distance also introduces instability in generation tasks.
On FLORES, CIF substantially improves Arabic and Hindi, but regresses on Chinese.
This indicates that while the centroid captures a useful shared multilingual representation, some distant languages may still contain language-specific projection information that is useful for generation.
This observation is consistent with our leave-one-language-out evaluation: removing the held-out language has almost no effect for Hindi and Vietnamese on FLORES, but hurts Chinese by $-2.35$ BLEU.
Overall, CIF is most reliable when the shared intervention representation dominates the language-specific deviation, while highly distant languages in open-ended generation may still benefit from retaining their own projection signal.

\subsection{Sensitivity to Intervention Strength and Trimming Size}
\label{sec:alpha_drop_sensitivity}

CIF has two inference-time hyperparameters: the intervention strength $\alpha$ and the trimming size $\mathrm{drop}$.
The intervention strength controls the magnitude of the fused operator, while $\mathrm{drop}$ controls how many projection matrices farthest from the centroid are removed before fusion.
We analyze both parameters to verify that the reported improvements are not caused by a single favorable hyperparameter setting.

All intervention methods use the same non-zero $\alpha$ grid,
$\{-1.0,-0.9,\ldots,-0.1,0.1,\ldots,0.9,1.0\}$.
For each dataset--backbone pair, we select one $\alpha$ according to the average validation performance across all target languages, and the same selected value is shared by every language in that setting.
Thus, no method uses language-specific intervention-strength tuning.
For CIF, we additionally tune $\mathrm{drop}\in\{0,1,2\}$ using the same validation criterion.
The setting $\mathrm{drop}=0$ corresponds to CIF without trimming.

Table~\ref{tab:drop_sensitivity} reports the effect of the trimming size.
The results show that the best trimming size varies across datasets.
For example, $\mathrm{drop}=1$ gives the best result on XCOPA, XNLI, and WMT23, while $\mathrm{drop}=2$ is slightly better on FLORES.
This confirms that trimming is an important part of CIF, but also that the optimal amount of trimming should be selected on validation data rather than fixed universally.

Table~\ref{tab:alpha_sensitivity} reports CIF performance under representative fixed $\alpha$ values.
The results show that performance changes noticeably with $\alpha$, especially for generation tasks.
For example, WMT23 improves from $32.44$ at $\alpha=-1.0$ to $42.24$ at $\alpha=0.4$, while FLORES improves from $28.46$ at $\alpha=-1.0$ to $38.57$ at $\alpha=-0.1$.
This confirms that intervention strength must be tuned fairly for all methods.
At the same time, CIF remains competitive across a broad range of non-zero $\alpha$ values, rather than depending on a single isolated setting.

\subsection{Model Checkpoints and Loading Details}
\label{app:model_checkpoints}

We use four HuggingFace checkpoints in all experiments: 
\texttt{Qwen/Qwen2.5-7B-Instruct}, 
\texttt{bigscience/bloomz-7b1-mt}, 
\texttt{meta-llama/Llama-3.1-8B-Instruct}, 
and \texttt{mistralai/Mistral-7B-Instruct-v0.3}. 
All checkpoints are used as released by their providers. 
We do not further pretrain the models. 
We do not fine-tune the models. 
We do not use LoRA or any other parameter-efficient adaptation method. 
We also do not use weight quantization.

The backbone model parameters are frozen in all experiments. 
CIF only learns offline projection matrices from parallel sentences. 
These projection matrices are then applied during inference as intervention operators. 
Therefore, the reported improvements come from inference-time intervention rather than parameter updates.

\begin{table*}[t!]
    \centering
    \scriptsize
    \begin{tabular}{llccccc}
        \toprule
        \textbf{Dataset} &
        \textbf{Checkpoint} &
        \textbf{Metric} &
        \textbf{Zero-shot} &
        \textbf{SADI} &
        \textbf{INCLINE} &
        \textbf{CIF} \\
        \midrule

        XCOPA & Qwen2.5-7B & F1 & 50.84 & 57.07 & 52.42 & 61.16 \\
        XCOPA & BLOOMZ-7B1-MT & F1 & 61.76 & 59.49 & 62.27 & 63.27 \\
        XCOPA & LLaMA-3.1-8B & F1 & 50.11 & 53.07 & 51.16 & 53.76 \\
        XCOPA & Mistral-7B & F1 & 49.98 & 50.56 & 50.07 & 51.24 \\
        \midrule

        XStoryCloze & Qwen2.5-7B & F1 & 84.96 & 83.16 & 81.42 & 85.01 \\
        XStoryCloze & BLOOMZ-7B1-MT & F1 & 77.92 & 78.78 & 80.76 & 83.04 \\
        XStoryCloze & LLaMA-3.1-8B & F1 & 83.35 & 75.00 & 82.94 & 83.71 \\
        XStoryCloze & Mistral-7B & F1 & 79.32 & 79.25 & 79.48 & 79.27 \\
        \midrule

        XWinograd & Qwen2.5-7B & F1 & 30.39 & 30.09 & 27.61 & 30.13 \\
        XWinograd & BLOOMZ-7B1-MT & F1 & 65.20 & 50.84 & 65.18 & 65.48 \\
        XWinograd & LLaMA-3.1-8B & F1 & 31.41 & 30.70 & 31.60 & 31.82 \\
        XWinograd & Mistral-7B & F1 & 47.43 & 46.68 & 63.01 & 67.50 \\
        \midrule

        XCSR & Qwen2.5-7B & F1 & 34.97 & 19.91 & 31.40 & 35.01 \\
        XCSR & BLOOMZ-7B1-MT & F1 & 46.60 & 42.89 & 46.54 & 47.31 \\
        XCSR & LLaMA-3.1-8B & F1 & 37.91 & 30.71 & 37.24 & 37.85 \\
        XCSR & Mistral-7B & F1 & 37.00 & 30.88 & 39.21 & 40.57 \\
        \midrule

        XNLI & Qwen2.5-7B & F1 & 43.57 & 49.40 & 45.84 & 49.57 \\
        XNLI & BLOOMZ-7B1-MT & F1 & 70.31 & 46.08 & 72.06 & 76.73 \\
        XNLI & LLaMA-3.1-8B & F1 & 62.72 & 50.75 & 64.45 & 64.81 \\
        XNLI & Mistral-7B & F1 & 47.63 & 50.64 & 51.74 & 50.19 \\
        \midrule

        MZSRE & Qwen2.5-7B & F1 & 36.50 & 36.42 & 36.29 & 36.72 \\
        MZSRE & BLOOMZ-7B1-MT & F1 & 45.50 & 29.18 & 45.57 & 46.09 \\
        MZSRE & LLaMA-3.1-8B & F1 & 39.19 & 38.75 & 39.07 & 39.17 \\
        MZSRE & Mistral-7B & F1 & 38.72 & 39.74 & 41.53 & 42.57 \\
        \midrule

        WMT23 & Qwen2.5-7B & BLEU & 39.48 & 40.73 & 45.06 & 47.69 \\
        WMT23 & BLOOMZ-7B1-MT & BLEU & 34.10 & 34.25 & 34.50 & 34.98 \\
        WMT23 & LLaMA-3.1-8B & BLEU & 41.33 & 40.72 & 45.27 & 44.74 \\
        WMT23 & Mistral-7B & BLEU & 43.38 & 42.23 & 43.25 & 43.38 \\
        \midrule

        FLORES & Qwen2.5-7B & BLEU & 37.21 & 36.24 & 36.48 & 38.31 \\
        FLORES & BLOOMZ-7B1-MT & BLEU & 51.71 & 49.95 & 52.60 & 52.87 \\
        FLORES & LLaMA-3.1-8B & BLEU & 41.15 & 42.99 & 42.82 & 45.50 \\
        FLORES & Mistral-7B & BLEU & 23.72 & 26.80 & 26.89 & 27.00 \\
        \bottomrule
    \end{tabular}
    \caption{
    Per-checkpoint results across all datasets.
    Classification tasks are reported as F1 percentage, while WMT23 and FLORES are reported as BLEU.
    The exact HuggingFace checkpoint names are provided in Appendix~\ref{app:model_checkpoints}.
    }
    \label{tab:per_checkpoint_results}
\end{table*}

Overall, CIF outperforms INCLINE on 29 out of 32 dataset--checkpoint pairs. 
The gains appear in both classification and generation tasks. 
This shows that CIF is not driven by a single model checkpoint or a single task type.

\subsection{Language Pairs Used for Centroid Construction}
\label{app:language_pairs}

For each dataset, we construct language-specific projection matrices from English--target-language parallel sentences. 
The fused CIF centroid is built from the target-language projections listed in Table~\ref{tab:language_pairs}. 
All projection directions map from the target language to English.

\begin{table*}[t!]
    \centering
    \scriptsize
    \begin{tabular}{lll}
        \toprule
        \textbf{Dataset} &
        \textbf{Target Languages} &
        \textbf{Language Pairs} \\
        \midrule
        XCOPA & et, id, it, sw, ta, th, tr, vi, zh & target $\rightarrow$ en \\
        XStoryCloze & ar, es, hi, id, ru, sw, zh & target $\rightarrow$ en \\
        XWinograd & fr, ja, pt, ru, zh & target $\rightarrow$ en \\
        XCSR & ar, de, es, fr, hi, it, ja, nl, pt, ru, sw, ur, vi, zh & target $\rightarrow$ en \\
        XNLI & ar, de, el, es, fr, hi, ru, sw, th, tr, ur, vi, zh & target $\rightarrow$ en \\
        MZSRE & de, es, fr, pt, ru, th, tr, vi, zh & target $\rightarrow$ en \\
        WMT23 & de, ja, ru, uk, zh & target $\rightarrow$ en \\
        FLORES & ar, el, es, fr, hi, ru, tr, vi, zh & target $\rightarrow$ en \\
        \bottomrule
    \end{tabular}
    \caption{
    Target languages used to construct the fused CIF centroid for each dataset.
    English is used as the reference language and is not counted as a target language.
    }
    \label{tab:language_pairs}
\end{table*}

\subsection{Multilingual Baseline Adaptation and Scalability}
\label{app:baseline_scalability}

We also clarify how pairwise baselines are adapted to the multilingual setting. 
INCLINE is a pairwise intervention method. 
It learns and applies one projection operator for each English--target-language pair. 
Thus, if there are $N$ target languages and $L$ transformer layers, INCLINE requires $N \times L$ deployed projection matrices at inference time.

CIF uses the same language-specific projections only during offline fusion. 
After fusion, CIF keeps one shared intervention operator for each layer. 
Thus, CIF only requires $L$ deployed projection matrices at inference time. 
This reduces the number of deployed operators by a factor of $N$ compared with pairwise deployment.

This difference is important for multilingual scaling. 
When more target languages are added, pairwise methods require one new set of projection matrices for each language. 
CIF instead absorbs the language-specific projections into a shared centroid operator. 
The final inference-time memory and deployment cost remain independent of the number of target languages used during fusion.

\subsection{Package Version}
\label{sec:package}
Pytorch version is 2.3.0 and python is 3.9.1.

\end{document}